\documentclass{article}

\usepackage{arxiv}

\usepackage[utf8]{inputenc} 
\usepackage[T1]{fontenc}    
\usepackage{hyperref}       
\usepackage{url}            
\usepackage{booktabs}       
\usepackage{amsfonts}       
\usepackage{nicefrac}       
\usepackage{microtype}      
\usepackage{lipsum}		
\usepackage{graphicx}
\usepackage{natbib}
\usepackage{doi}

\usepackage[T1]{fontenc}
\usepackage{float}
\usepackage{amsmath}
\usepackage{mathtools}
\usepackage{graphicx}
\usepackage{multicol}
\usepackage{rotating}
\usepackage{multirow}
\usepackage{subcaption}
\usepackage{algorithm,algpseudocode}

\makeatletter
\def\algbackskip{\hskip-\ALG@thistlm}
\makeatother

\title{LIMEcraft: Handcrafted superpixel selection and inspection for Visual eXplanations}

\date{} 					

\author{ \href{https://orcid.org/0000-0003-2903-6050}{\includegraphics[scale=0.06]{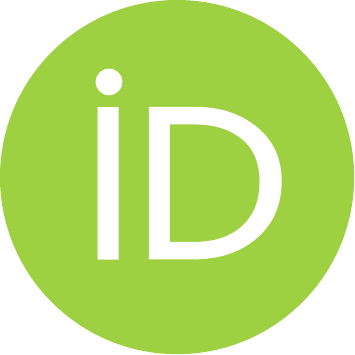}\hspace{1mm}Weronika Hryniewska} \\
	Faculty of Mathematics and Information Science\\
	Warsaw University of Technology\\
	Koszykowa 75, 00-662 Warsaw (Poland) \\
	\texttt{weronika.hryniewska.dokt@pw.edu.pl} \\
	\And
    \hspace{1mm}Adrianna Grudzień \\
	Faculty of Mathematics and Information Science\\
	Warsaw University of Technology\\
	Koszykowa 75, 00-662 Warsaw (Poland) \\
	\texttt{a.grudzien@student.mini.pw.edu.pl} \\
	\And
	\href{https://orcid.org/0000-0001-8423-1823}{\includegraphics[scale=0.06]{orcid.pdf}\hspace{1mm}Przemysław Biecek} \\
	Faculty of Mathematics and Information Science\\
	Warsaw University of Technology\\
	Koszykowa 75, 00-662 Warsaw (Poland) \\
	\texttt{przemyslaw.biecek@pw.edu.pl} \\
}

\renewcommand{\shorttitle}{LIMEcraft}

\hypersetup{
pdftitle={LIMEcraft},
pdfsubject={q-bio.NC, q-bio.QM},
pdfauthor={Weronika Hryniewska, Adrianna Grudzień, Przemysław Biecek},
pdfkeywords={Explainable AI, superpixels, LIME, image features, interactive User Interface},
}

\begin{document}
\maketitle

\begin{abstract}
    The increased interest in deep learning applications, and their hard-to-detect biases result in the~need to validate and explain complex models. However, current explanation methods are limited as far as both the~explanation of the~reasoning process and prediction results are concerned. They usually only show the~location in the~image that was important for model prediction. The~lack of possibility to interact with explanations makes it difficult to verify and understand exactly how the~model works. This creates a~significant risk when using the~model. The risk is compounded by the~fact that explanations do not take into account the~semantic meaning of the~explained objects. To escape from the~trap of static and meaningless explanations, we propose a~tool and a~process called LIMEcraft. LIMEcraft enhances the~process of explanation by allowing a~user to interactively select semantically consistent areas and thoroughly examine the~prediction for the~image instance in case of many image features. Experiments on several models show that our tool improves model safety by inspecting model fairness for image pieces that may indicate model bias. The~code is available at: \url{http://github.com/MI2DataLab/LIMEcraft}
\end{abstract}


\begin{figure}[!h]
    \centering
    \includegraphics[width=\textwidth]{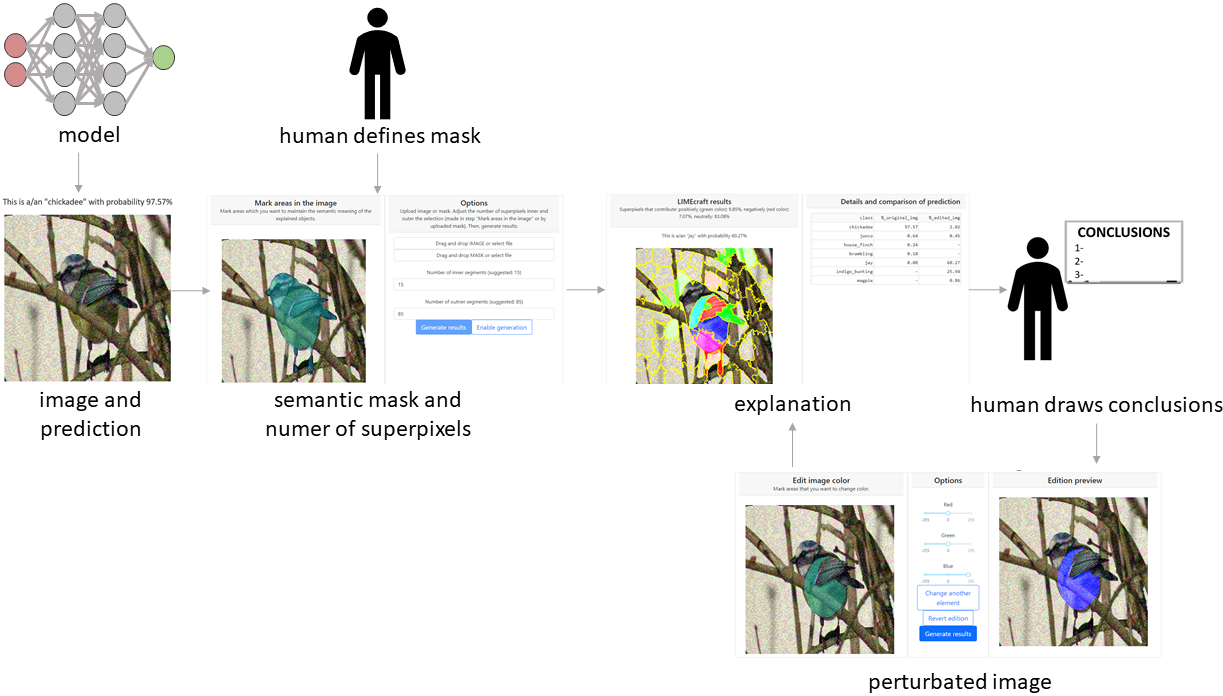}
    \caption{Diagram summarizing the~explanation process using LIMEcraft}
    \label{fig:process}
\end{figure}

\section{Introduction}\label{s1}

Artificial Intelligence (AI) is rapidly becoming applicable in a~variety of domains. Deep learning (DL) has already achieved significant results in many areas concerning computer vision. However, despite these remarkable results achieved by DL, the~decisions made by black-boxes still remain unclear for humans, due to difficulty in understanding the~reasoning process of the~neural network.The lack of interpretability results in critical issues considering model fairness and safety.

For this reason, explainability methods have begun to attract researchers' attention. They have started to create various approaches to explain neural networks' decision process. One vastly used method is Local Interpretable Model-Agnostic Explanations (LIME). It appears that explanation which marks important regions in the~image is easily understandable for humans, and therefore used in many scientific studies.

However, so far, most of these methods consider it sufficient to mark only the~area that affects the~model prediction. The~construction of explanations does not take into account the~individual factors that contribute to the~significance of a~region in the~model prediction.

In this paper, we propose a~new process of explanation based on LIME with the~possibility of inspection image features, such as: color, shape, position, and rotation for creation of Visual eXplanations. LIMEcraft also allows handcrafted superpixel selection, which eliminates non-interaction problems with explanation methods and improves the~explanation quality of complex image instances. The human interaction process is described in Figure \ref{fig:process}.

\section{Related works}
\subsection{Perturbation-based explainable algorithms}

Perturbation-based explainable algorithms use a~technique of iteratively removing or changing parts of image features. The~variety of perturbations shows how many types of distortions can be applied to images. Among such techniques, we can distinguish: occlusion (e.g. LIME by \cite{Ribeiro2016} or RISE by \cite{Petsiuk2018rise}), DeConvolution and Occlusion Sensitivity  (see \cite{Zeiler2014}), blurring \citep{Fong2017}, conditional sampling (Prediction Difference Analysis by \cite{Zintgraf2017}), adding noise (Noise Sensitivity by \cite{greydanus18a}), substitution of existing features (IRT and OSFT by \cite{Burns2019}) or superimposing another image \citep{AnchorLIME}. Based on a~model response, the~importance of those image features is calculated. Then, the~attribution of each feature is computed and the~results are shown. However, such techniques do not measure the~importance of particular image features, only the~location of parts of the~image is important. The~lack of investigation which image features (color, position, shape, brightness) play the~most crucial role, makes these methods prone to errors.

One of the~most popular techniques that use occlusion to check the~importance of regions in an~image is called Local Interpretable Model-Agnostic Explanations \cite[LIME]{Ribeiro2016}). LIME is a~model-agnostic explanation algorithm. Model-agnostic means that the~architecture of the~model does not have an~influence on the~possibility to explain the~model. The~explanation is local - it focuses on one specific prediction rather than considering the~model globally. The~LIME algorithm works with tabular data, text, and images. As for images, it divides them into superpixels based on the~quick shift algorithm \citep{Vedaldi2008}. Quick shift is a~fast mode-seeking algorithm that segments an~image by localizing clusters of pixels in both spatial and color dimensions. Then, a~dataset with some superpixels occluded is generated. Each perturbed instance gets the~probability of belonging to a~class. On this locally weighted dataset, the~linear model is trained. The~highest positive and negative weights for a~specific class are presented in the~original image by addition, respectively, a~green or a~red semitransparent mask on the~most important superpixels.

\subsection{Methods based on LIME}

Following the~success of LIME algorithm, many scientists started to be interested in developing this method to make it even more efficient and effective. The~greatest number of LIME modifications are for tabular data, e.g., DLIME \citep{dlime}, GraphLIME \citep{graphlime}, Tree-LIME \citep{tree-LIME}, ALIME \citep{alime}, LIME-SUP \citep{limesup}. However, methods based on LIME for the~images are also developed and published in scientific journals, namely: Anchor LIME \citep{AnchorLIME}, LIMEAleph \citep{LIME-Aleph}, KL-LIME \citep{kl-lime}, MPS-LIME \citep{MPSLIME}, and NormLIME \citep{NormLIME}.

Anchor LIME \citep{AnchorLIME}, instead of hiding some superpixels from the~original image, superimposes another image over the~rest of the~superpixels. Authors stress that the~method might seem unnatural, but it allows to predict the~model's behavior on unseen cases.

A~LIME-based approach Kullback Leibler divergence, called KL-LIME \citep{kl-lime}, is designed for~explaining the~predictions of Bayesian predictive models. The~proposed method combines methods from Bayesian projection predictive variable selection with LIME algorithm. In KL-LIME, parameters of the~interpretable model are found by minimizing the~Kullback–Leibler divergence from the~predictive model.

LIME-Aleph \citep{LIME-Aleph} combines an~explanation generated by LIME with logic rules obtained by the~Inductive Logic Programming system Aleph. The~authors claim that in LIME it is not clear if the~classification decision is made due to the~presence of specific parts of the~image or because of the~specific relation between them. Their approach is capable of identifying the~relationship between elements as an~important explanatory factor.

The method of superpixels selection is replaced in MPS-LIME \citep{MPSLIME} with Modified Perturbed Sampling (MPS) operation. MPS-LIME converts superpixels into an~undirected graph. The~authors claim that their method does not ignore the~complicated correlation between image features and improves the~algorithm efficiency.

LIME is a~method for a~local explanation, while NormLIME \citep{NormLIME} tries to aggregate local explanations and create a~global, class-specific explanation.

\subsection{Limitations of existing algorithms}

There are some weaknesses of the~LIME method. The~definition of superpixels does not take into account the~semantic meaning of objects in the~image, and consequently, sometimes different objects are located within a~single superpixel. This is especially visible in images with many overlapping objects and in medical images.

Moreover, despite many attempts to improve the~LIME algorithm, it is often considered non-robust. \cite{Alvarez-Melis2018} show that perturbation-based methods are especially prone to instability. Small changes in the~input image, such as adding Gaussian noise, should not significantly affect explanations. However, due to the~fully automatic selection of the~superpixels, LIME depends strongly on nonsemantic input image features and is particularly sensitive to noise.

A suggestion that the~existing explanation techniques are vulnerable to attacks of adversarial classifiers is made by \cite{Slack2020Adversarial}. They claim that LIME is not sufficient for ascertaining the~discriminatory behavior of classifiers in sensitive applications and is not reliable. Their approach can be used to scaffold a~biased classifier. Predictions of the~classifier on the~input data still remain biased, but the~post hoc explanations of the~scaffolded classifier look innocuous.

Moreover, \cite{Rahnama2019} claim that LIME suffers from data and label shift. Their experiments show that the~instances generated by LIME’s algorithm are distinctly different from training instances drawn from the~underlying distribution. Based on the~obtained results, they conclude that random perturbations of the~features of the~explained instance cannot be considered a~reliable method of generating data in the~LIME method.

\cite{Schallner2020} stress that the~selection of a~suitable superpixel algorithm should be considered. They conduct several experiments comparing different superpixel algorithms. Finally, they say that for each problem, the~superpixel selection algorithm should be consulted with domain experts. In some situations, it is important to generate superpixels of significantly different sizes depending on the~semantic meaning of the~content. 

To sum up, LIME is sensitive to small changes in the~input image, such as adding noise or an~adversarial attack. Random feature perturbations are unreliable because LIME suffers from data and label shift. The~choice of an~optimal segmentation algorithm affects the~LIME results and should be consulted with a~domain expert. As we will show below, for partially covered or noisy features, automatic segmentation leads to worse results than expert-assisted segmentation. 

\section{LIMEcraft}

\subsection{Motivation}

During our work with medical images, we discovered that LIME prefers sharp boundaries between superpixels. It also needed further clarification on which kind of image features the~prediction of a~specific part of the~image was made. Our motivation was to create a~solution that not only works better on LIME's corner cases, but also gives more detailed insight into the~model's strengths and weaknesses.

\begin{figure}[ht]
    \centering
    \includegraphics[width=3.8cm]{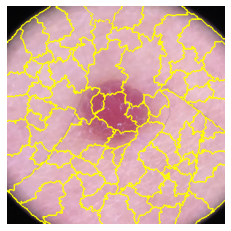}
    \includegraphics[width=3.8cm]{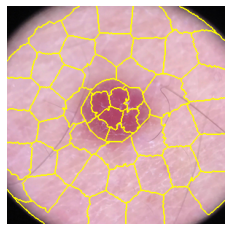}
    \hspace{0.3cm}
    \includegraphics[width=3.8cm]{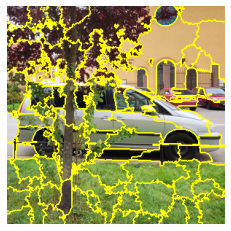}
    \includegraphics[width=3.8cm]{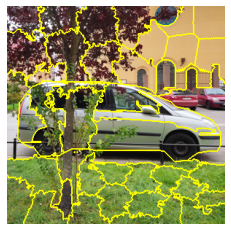}
    \caption{Examples of division into superpixels by LIME and by our algorithm of skin lesion~\citep{Vargas2019} and car photo. For each pair, the~left image shows the~automatic LIME segmentation and the~right one a~LIMEcraft segmentation supported by the~user who outlined the~skin lesion of interest (first pair) or the~car (second pair). LIME uses quick shift algorithm, and LIMEcraft uses manual or predefined superpixel selection, and then, segmentation based on K-means clustering algorithm.}
    \label{fig:strangeLIMEexamples}
\end{figure}

The weakness of LIME is that the~division into superpixels is unacceptable when objects are partially covered or the~image is noisy. LIME also fails when there are many elements that interfere with the~way superpixels are constructed, e.g. zebra stripes. The lack of semantic understanding of the~image makes LIME unsafe for many medical images and for images taken in the~natural environment when some objects are partially obscured by others, as presented in Figure~\ref{fig:strangeLIMEexamples}. For this reason, we create an~opportunity to manually select superpixels.

The next problem that we wanted to address is the~lack of understanding of the~image features that mostly contribute to the~model prediction. The~location of the~most important superpixel does not provide us with complete information about whether the~model has learned the~correct features. Without careful verification, we cannot be sure that a~car of any color will be correctly recognized by the~model as a~car and classified into the~correct class. Image features' inspection may also help us to investigate the~possibility of bias based on, e.g., skin color.

The LIME algorithm is prone to the~presence of noise. Noise can greatly interfere with how superpixels are formed, and thus, due to the~large differences in superpixel sizes, also change the~areas that are marked by LIME as relevant to the~model. Although our solution cannot fully eliminate this instability, it limits an~uncontrolled splitting into superpixels and improves the~reliability testing capabilities of the~model.

In summary, LIME's weaknesses are: (1) uninformative/misleading segmentation, due to the~lack of semantic understanding of the~image during the~creation of superpixels, (2) lack of understanding of the~image features that mostly contributed to the~model prediction, (3) and sensitivity to the~presence of noise.

\subsection{Different types of superpixels' selection}

In~LIMEcraft there are two main ways of selecting segments called “superpixels”: semantic and non-semantic. The first one can be done by a~user using a~tool, which allows drawing an~irregular path of shape, or by uploading a~prepared mask of superpixels. Such functionalities lead to greater influence on proper image analysis. 

After manual or predefined superpixel selection, the~next step is the~non-semantic selection of superpixels. In this step, only previously selected areas are divided into smaller pieces. Such automatic segments are generated using image segmentation based on the~K-means clustering algorithm. Moreover, we can determine into how many segments the~areas selected will be divided. LIMEcraft suggests how many superpixels will be optimal for each case. It calculates how many superpixels should be inside and outside the~selected areas to maintain the~same size of the~superpixels. However, in some cases, the~user may want to increase the~number of superpixels inside the~selected areas to obtain more detailed results. It might be helpful when the~object inside the~selection has small details and the~rest of the~image is just a~little diverse background.

The potential use-case scenario is to manually define objects that should not be combined in the~same superpixel. Such an~image could be, for example, a~complex cityscape partially obscured by tree branches, an~X-ray image in which there are naturally small differences in the~brightness of areas while these areas belong to other internal organs, a~photograph of undergrowth in which there are many objects similar in color.

Checking the~responsibility of a~network to classify lung lesions can be an~example of using the~mask loading functionality. The~neural network was trained on the~data with lesions label, and then validated on an~external database, as recommended by \cite{Hryniewska2021}. For external validation, the~dataset for lesions detection was chosen, so the~database contained not only the~names of lesions present on the~images but also their location. The masks of lesions can be easily uploaded into LIMEcraft to verify whether they are important for the~model's prediction. The dashboard shows how much of the~whole image has a~positive impact on the~prediction of the~model (green color), negative (red color), and neutral (without color). It might be useful to assess the~importance of the~selected areas (when the~superpixels are not exactly the~same size).

\subsection{Inspection of image feature importance}

The interface we have created makes it possible to analyze the~impact of image perturbation on the~prediction of the~model. The~user can edit the~color, shape, and position of the~selected area, and then subject the~edited image to the~LIMEcraft algorithm. 

Color edition enables to manipulate the~values of individual channels of the~image (RGB), so the~brightness can also be adjusted. Moreover, we can rotate the~selected area and change its position. The~moved piece can also be completely removed. In the~edited area, the~inpainting algorithm based on “biharmonic equation” \citep{Damelin2018} is applied. The~selected area may also be expanded according to a~user-defined “power” value. For values greater than 1, it will be enlarged, and for values less than 1, it will be shrunk.

In order to be able to better understand the~changes that have occurred as a~result of the~perturbation of the~image, the~report in a~form of a~table is generated. It compares the~percentages of probabilities for the~predicted classes.

It is crucial to see how the~model responds to a~change in individual image elements. For example, for microscope images, the~shapes and colors of cells may be relevant to a~classification task. However, position and rotation should not significantly affect the~prediction of the~trained model. By running such experiments using LIMEcraft, we can observe whether the~model has learned to recognize objects by the~correct features.

\begin{figure}[hb]
    \centering
    \includegraphics[width=\linewidth]{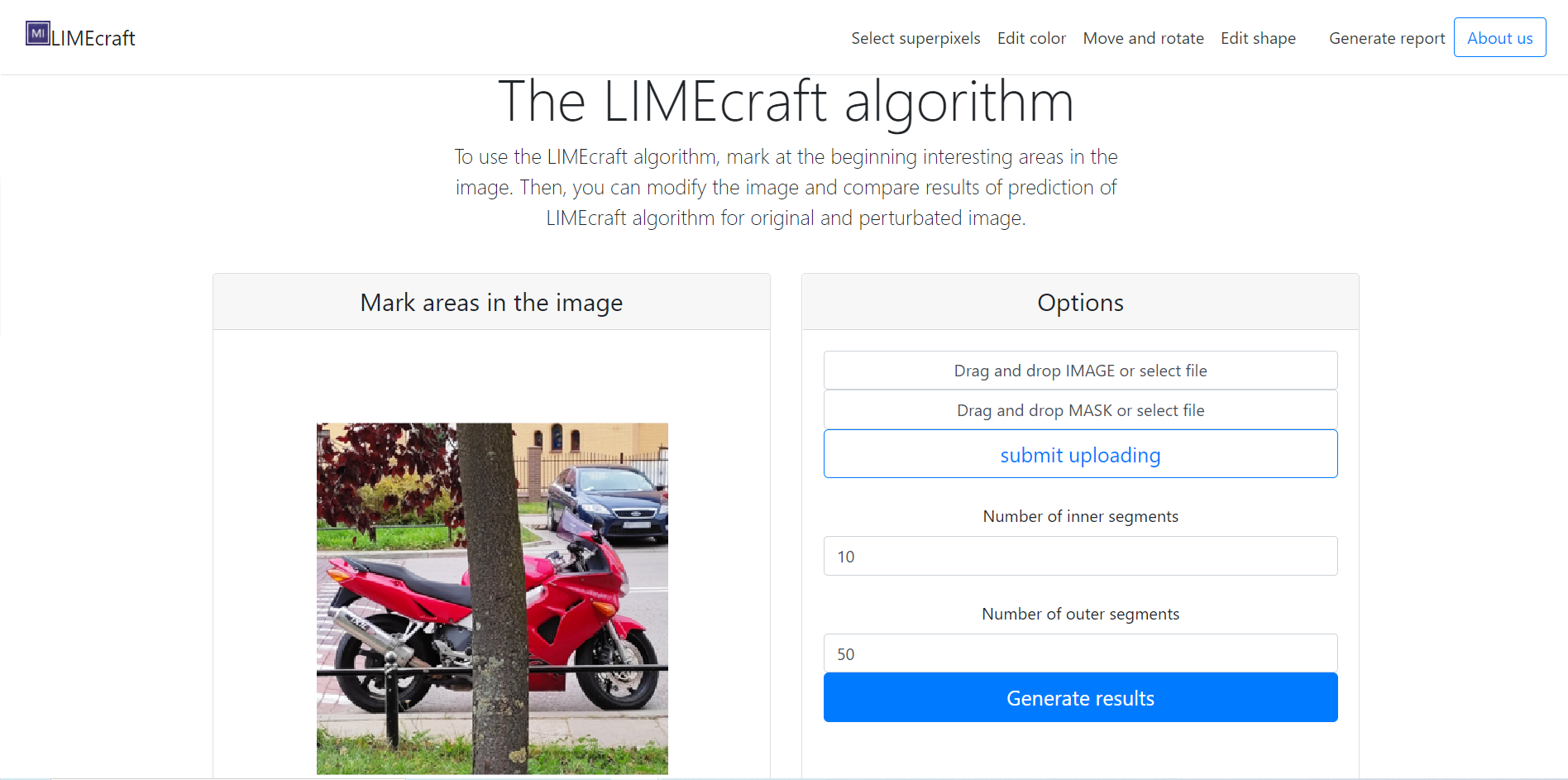}
    \caption{An example of the~user interface for code available on \url{http://github.com/MI2DataLab/LIMEcraft} allowing to outline features of interest and analyze the~explanations yourself. }
    \label{fig:2dinterface}
\end{figure}

\subsection{Interactive User Interface}

In contrast to the~fully automated approach of the~LIME algorithm, our LIMEcraft algorithm incorporates the~human into the~process of explainability. It gives them the~ability to influence the~division into superpixels, the~choice of the~number of superpixels, and a~more detailed analysis of the~model by comparing the~prediction results for the~original image and the~one subjected to perturbations.

The undeniable advantage of the~dashboard is that it can be used by people unfamiliar with programming, because the~interface is very intuitive and user-friendly. The~interface is presented in Figure \ref{fig:2dinterface}.

The interactive User Interface gives the~human more control over the~quality of the~model and, as a~result, the~safety of the~created models. An~important aspect that cannot be ignored is the~variety of biases that a~model may have. With an~interactive interface, different possibilities (image modifications) can be tested to ensure safety and fairness. A~good example to consider here would be the~possibility of changing the~skin color of a~person in a~photo.

\subsection{Algorithm in details}

LIMEcraft algorithm is based on LIME. However, there are some key differences between both of them. As presented in Algorithm \ref{limecraft}, LIMEcraft algorithm includes the~possibility to define mask for the~selected image, and then, to choose the~number of superpixels inside and outside the~mask. The~next important innovation is a~functionality to edit image, which provides the~insight into the~model's robustness. Moreover, LIME and LIMEcraft have different segmentation algorithms. LIME uses quick shift algorithm. In LIMEcraft, besides manual or predefined superpixel selection, segmentation is based on K-means clustering algorithm.

\begin{algorithm}[H]
    \caption{LIMEcraft algorithm}\label{limecraft}
    \hspace*{\algorithmicindent} \textbf{Input:} black-box model $f$, input sample $image$, mask of superpixels $M$, number of superpixels $n_1$ (inner of selected area), $n_2$ (outer of selected area), number of features to pick $m$ \\
    \hspace*{\algorithmicindent} \textbf{Output:} explainable coefficients from the~linear model
    \begin{algorithmic}[1]
        \If{you want} \;\;\;\;\;\;\;\;\;\;\;\;\; $\vartriangleright Define\: mask $
            \State $mask \gets UploadMask(M)$
        \Else
            \State $mask \gets MarkAreasInTheImage$
        \EndIf
         \If{you want} \;\;\;\;\;\;\;\;\;\;\;\;\; $\vartriangleright Choose\: number\: of\: superpixels $
        \State $ x_1 \gets segment.KmeansClustering(mask, n_1)$
        \State $ x_2 \gets segment.KmeansClustering(image-mask, n_2)$
        \State $ x \gets x_1 + x_2$
        \EndIf
        \If{you want}  \;\;\;\;\;\;\;\;\;\;\;\;\; $\vartriangleright Change\: image\: features $
            \State Edit image
        \EndIf
        
        \State $\overline{y} \gets f.predict(x) $
        
        \For {$\textit{i}$ in $\textit{n}$} \do{}
            \State $p_i \gets Permute(x) \;\;\;\;\;\;\;\;\; \vartriangleright Randomly\: pick\: superpixels $
            \State $obs_i \gets f.predict(p)$
            \State $dist_i \gets \mid \overline{y}-obs_i\mid $
        \EndFor
        
        \State $simscore \gets SimilarityScore(dist)$
        \State $x_{pick} \gets Pick(p, simscore, m)$
        \State $L \gets LinearModel.fit(p, m, simscore)$
        \State \textbf{return} \emph{L.weights}
    \end{algorithmic}
\end{algorithm}

\section{Method evaluation}

\subsection{Relevance of image features to correctness of prediction}

\begin{figure}[h]
    \centering
    \subcaptionbox{Skin  lesion}{\includegraphics[width=3.5cm]{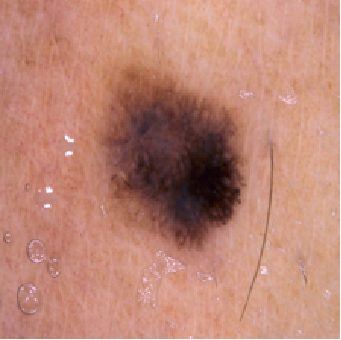}}\hfill
    \subcaptionbox{Selected mask}{\includegraphics[width=3.5cm]{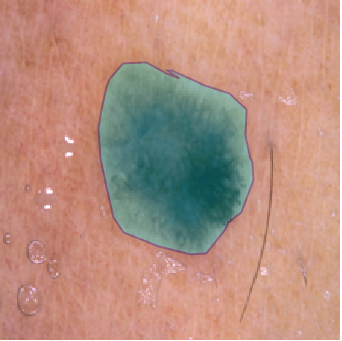}}\hfill
    \subcaptionbox{LIMEcraft results melanocytic nevi 54.79\%}{\includegraphics[width=3.5cm]{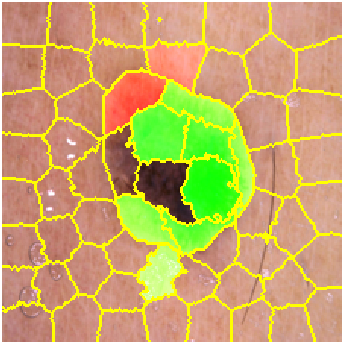}}\hfill
    \subcaptionbox{LIME results}{\includegraphics[width=3.5cm]{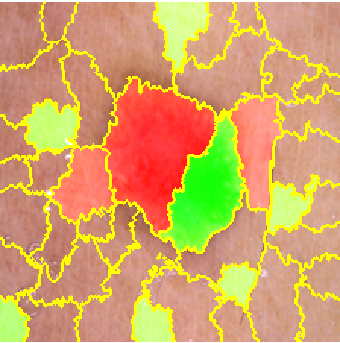}}\\
    \subcaptionbox{Color edition}{\includegraphics[width=3.5cm]{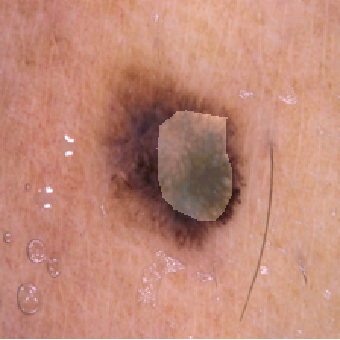}}\hfill
    \subcaptionbox{LIMEcraft for (e) benign keratosis-like 99.80\%}{\includegraphics[width=3.5cm]{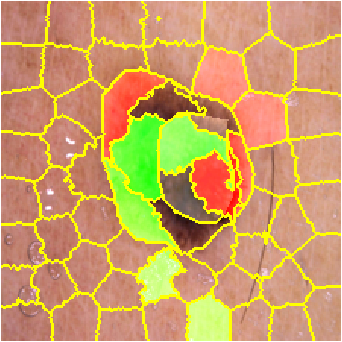}}\hfill
    \subcaptionbox{Shape edition}{\includegraphics[width=3.5cm]{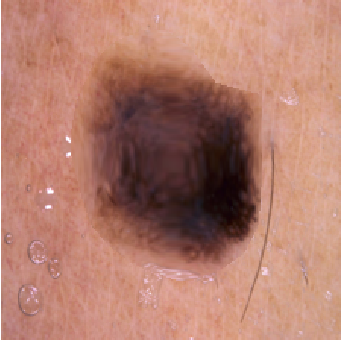}}\hfill
    \subcaptionbox{LIMEcraft for (g) melanoma 52.86\%}{\includegraphics[width=3.5cm]{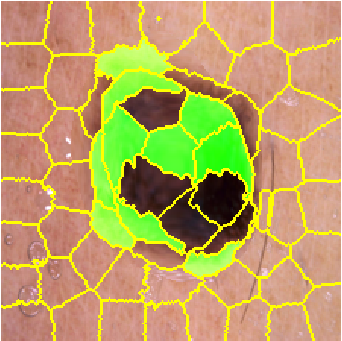}}\hfill
    \\
     \subcaptionbox{Hair removal}{\includegraphics[width=3.5cm]{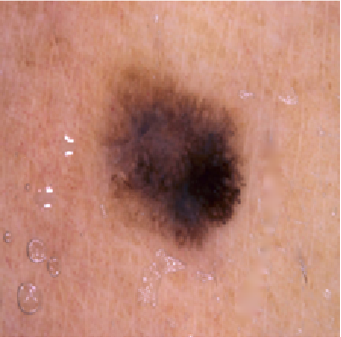}}\hfill
    \subcaptionbox{LIMEcraft for (i) melanoma 66.36\%}{\includegraphics[width=3.5cm]{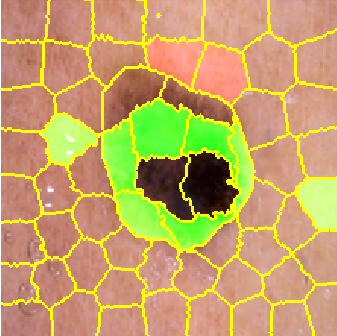}}\hfill
    \subcaptionbox{Color, rotation and location edition}{\includegraphics[width=3.5cm]{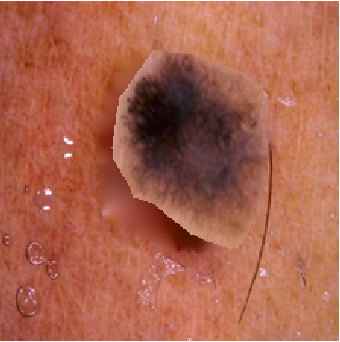}}\hfill
    \subcaptionbox{LIMEcraft for (k) benign keratosis-like 98.82\%}{\includegraphics[width=3.5cm]{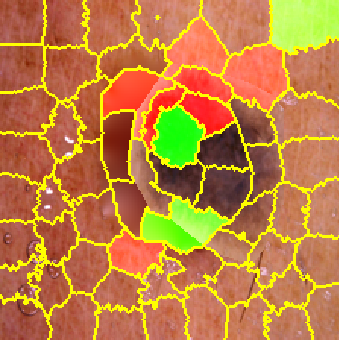}}\hfill
    \caption{Model inference results for skin lesion classification using the~LIMEcraft algorithm. The~input image shows a~lesion called melanocytic nevi. Superpixels colored to green mean that this part of the~image contributes positively to the~prediction, while red parts show negative impact.} 
    \label{fig:whole_experiment}
\end{figure}

We test the~ability to evaluate a~model for skin lesion classification using LIMEcraft. First of all, we select a~database for classification of skin lesions \citep{Mader2019}. Then, we choose model architecture: MobileNet and image input size: 224x224x3. The~neural network uses the~base pretrained on Imagenet. While classifying into 7 classes, it achieves 64.6\% of sparse categorical accuracy.

To investigate if the~model is predicting class label based on skin lesions, not artifacts, such as hair, we conduct several tests using LIMEcraft. We select a~mask (presented in Figure \ref{fig:whole_experiment} (b)), and we run several experiments of feature importance. 

The color edition, visible in Figure \ref{fig:whole_experiment} (e), changes the~model prediction from class melanocytic nevis (54.79\%) to benign keratosis-like (99.80\%). 

LIMEcraft results obtained after shape edition (power of expansion: 1.4), in Figure \ref{fig:whole_experiment} (g), do not change model's confidence so drastically, because it drops only 8.5\%. However, it changes the~most probable class to melanoma with 52.86\% of probability.

In Figure \ref{fig:whole_experiment} (i)), we remove hair and the~prediction of class melanocytic nevi decreases to 33.54\%. The~most probable class for this case is melanoma. 

The last conducted experiment covers many different image editions, namely: 10px shift into right and down direction, 180$^{\circ}$ rotation, and change of patient's skin color.

Based on \citep{Stieler_2021_CVPR}'s work, it can be presumed that changes such as rotation and shift of the~skin lesion should not change the~prediction of the~model. On the~other hand, changing the~color and boundaries may change the~features of the~lesion.

\begin{figure}[H]
    \centering
    \includegraphics[width=12cm]{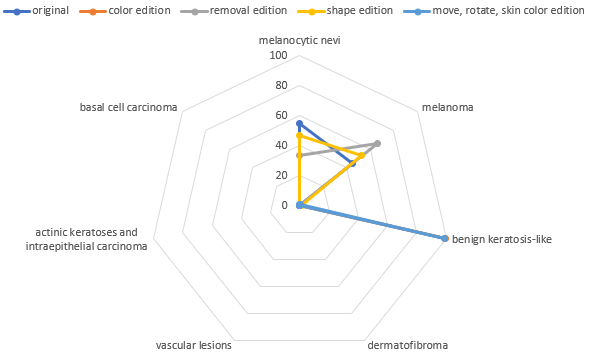}
    \caption{Radar plot of the~model's confidence in class selection for a~skin lesion called melanocytic nevi.}
    \label{fig:evaluation_schema}
\end{figure}

Figure \ref{fig:evaluation_schema} partially confirms our assumptions. The~color of the~lesion had a~large effect on the~prediction of the~model. Hair removal does not drastically change the~model's prediction. Nevertheless, there are some disturbing findings. A~combination of several changes: skin color, shape, and rotation had a~great influence on model classification result. This result may become the~basis for speculating that the~model could be biased.

Nevertheless, it is very important to note that results obtained by using the~LIMEcraft algorithm should be evaluated by a~domain specialist. It is up to the~specialist to determine whether a~given change in the~input image should cause a~change in the~model response.

\subsection{Noise example-based sensitivity analysis}

\begin{figure}
    \centering

    \def\arraystretch{1.4}
    \setlength{\tabcolsep}{0.6pt}
    
    \begin{tabular}{cccccc}
                                   &
    No perturbation                &
    $\sigma$=0.2                   &
    $\sigma$=0.4                   &
    $\sigma$=0.6                   &
    $\sigma$=0.8                   \\
    \multirow{1}{*}[14ex]{\rotatebox[origin=c]{90}{Original image}} &
    \includegraphics[width=3.1cm]{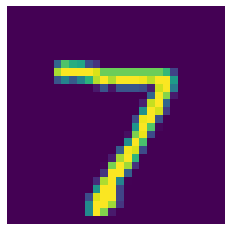} &
    \includegraphics[width=3.1cm]{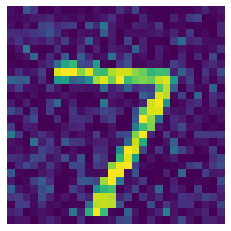} &
    \includegraphics[width=3.1cm]{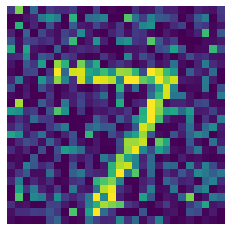} &
    \includegraphics[width=3.1cm]{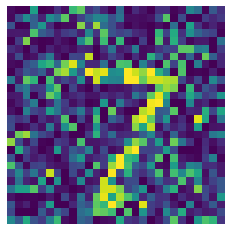} &
    \includegraphics[width=3.1cm]{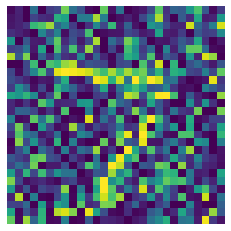} \\
    \multirow{1}{*}[14ex]{\rotatebox[origin=c]{90}{LIME mask}} &
    \includegraphics[width=3.1cm]{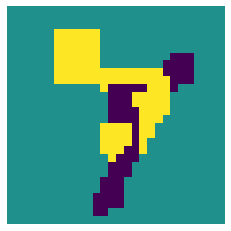} & 
    \includegraphics[width=3.1cm]{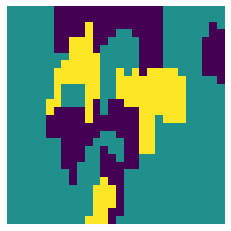} &  
    \includegraphics[width=3.1cm]{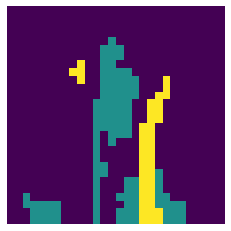} &
    \includegraphics[width=3.1cm]{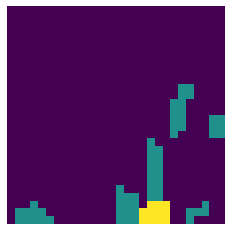} &
    \includegraphics[width=3.1cm]{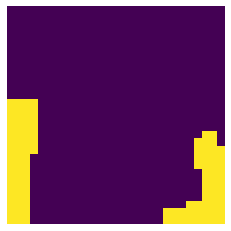} \\
    \multirow{1}{*}[16ex]{\rotatebox[origin=c]{90}{LIMEcraft mask}} & \includegraphics[width=3.1cm]{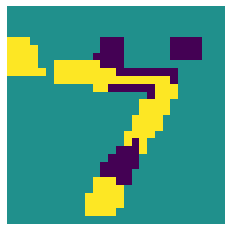} & \includegraphics[width=3.1cm]{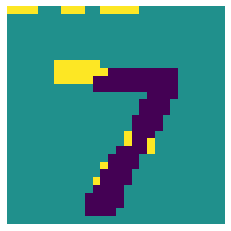} & \includegraphics[width=3.1cm]{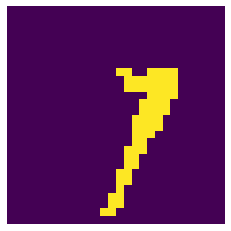} &
    \includegraphics[width=3.1cm]{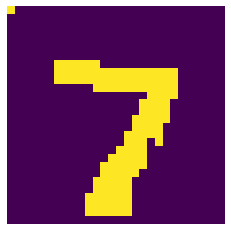} &
    \includegraphics[width=3.1cm]{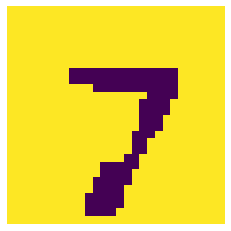} \\
    \end{tabular}

    \caption{Images of digit 7 perturbed with Gaussian noise of $\sigma$ strength and images showing the~significance of superpixels on the~classification of digit 7 ($-1$ negatively affects prediction, 0 has no significant effect, 1 positively affects) generated by the~LIME and LIMEcraft algorithms.}
    \label{fig:sensitivity_analysis}
\end{figure}

\cite{Slack2020Adversarial} state that LIME can be easily manipulated to hide biases. It stresses the~fact that such explanations are untrustworthy and not safe in usage.

We want to examine if LIMEcraft improves the~robustness of model explanations. \cite{Alvarez-Melis2018} add Gaussian noise to the~input image and check how strongly the~output generated by explanation varies. They show that the~LIME explanation is highly vulnerable to small changes in input image due to the~presence of sparse superpixels.

To test LIMEcraft safety, we train on MNIST dataset small neural network with only one convolutional layer with relu activation, max-pooling, flattening, and one dense layer. It receives 98.07\% accuracy on the~test set. 

Then, we perturb the~image with number 7 by adding Gaussian noise that varies in intensity. The~input image has pixel intensities between [0, 1] and the~values out of this range are clipped. In Figure \ref{fig:sensitivity_analysis} saved masks of superpixels generated by both LIME and LIMEcraft are presented. In order to obtain more comparable evaluation results for both algorithms, the~same segmentation algorithm is chosen - based on K-means clustering. 

\begin{figure}[ht]
    \centering
\includegraphics[width=\linewidth]{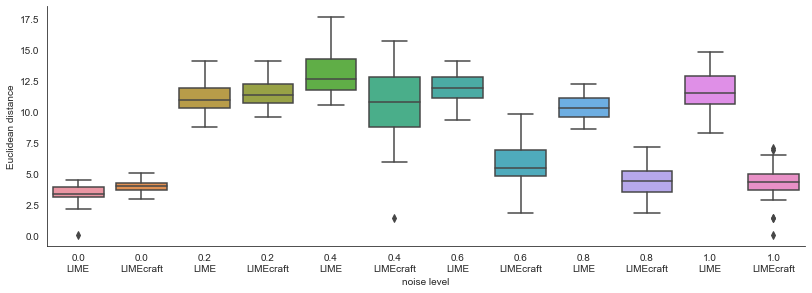}
    \caption{Box plot of the~noise level dependence of the~10 images from the~MNIST database on the~similarity of the~masks generated by LIMEcraft and LIME computed by Euclidean distance for each pair of images.}
    \label{fig:boxplot}
\end{figure}

In the~experiment presented in Figure \ref{fig:boxplot}, it appears that, thanks to the~manual selection of superpixel covering the~number, LIMEcraft is more stable in case of the~presence of noise. The~Euclidean distance for strongly noisy images is not as high as for LIME algorithm. Not changing the~outer border of the~7 number in LIMEcraft leads to an~improvement in robustness to perturbations in the~input. It is also worth noting that for no noise or low noise values, the~masks generated by LIMEcraft are comparable to the~results achieved by LIME. Thus, it can be concluded that LIMEcraft improves the~repeatability of the~obtained explanations.

\subsection{Evaluation with human subjects}
The user study consisted of a~pilot user study, and then a~formal user study of 20 people. After the~pilot study, the~remarks reported by users were used to improve LIMEcraft and any ambiguities in the~survey were clarified before proceeding with the~formal user study.

The aim of the~participants was to follow the~instructions to complete a~task and fill a~questionnaire. Participants tested ImageNet model of Inception v3 architecture with 5 different non-medical images available in Github repository of LIMEcraft: \url{https://github.com/MI2DataLab/LIMEcraft}. They had to assess a~model quality by seeing what a~model has learned and to check the~model's safety using the~LIME and LIMEcraft explanations. Using a~questionnaire, we wanted to assess whether people think LIMEcraft enhances their explanatory abilities and what features of LIMEcraf are most important to them. 

\begin{figure}[!h]
    \centering
\includegraphics[width=\linewidth]{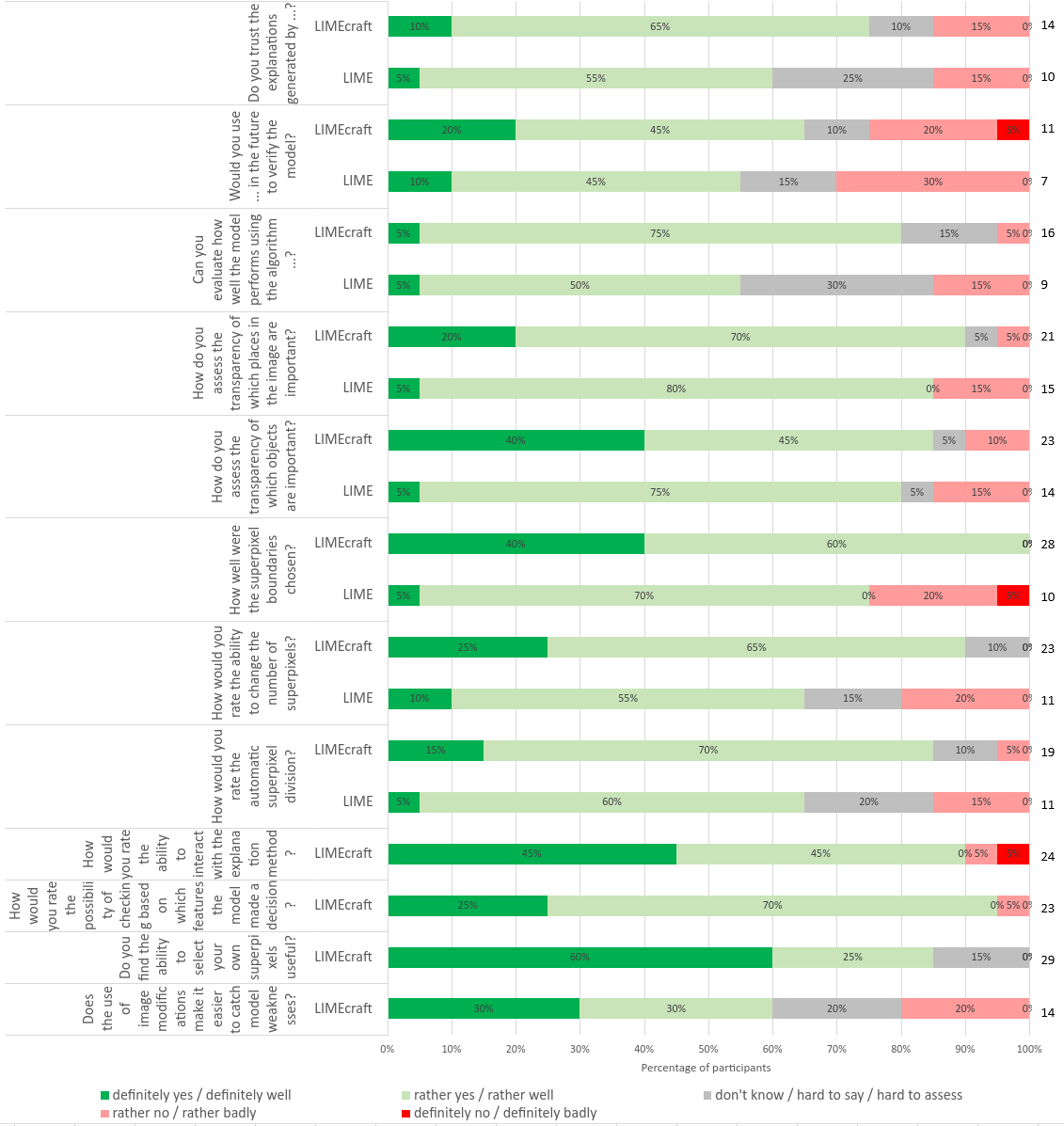}
    \caption{Stacked column chart showing the~responses of 20 users to a~survey designed to evaluate and compare LIMEcraft and LIME. Each row is related to a~separate question. The colors encode the~respondents' answers and are then converted to numbers according to the~legend. On the~right side of the~chart is the~sum of the~ratings of all respondents.}
    \label{fig:stackedcolumnchart}
\end{figure}

In the~formal user study, 15 men and 5 women of varying age participated. No strong correlation was discovered between sex and responses. Users' experience in using AI models and LIME differed. The AI experience declared by users had a~Gaussian-like distribution. 40\% of the~people did not know LIME, and the~remaining number of users was roughly equally distributed between those who knew the~method poorly, moderately, and well. The median time required to participate in the~experiment and respond was 38 minutes and the~interquartile range was 34 minutes.

In Figure \ref{fig:stackedcolumnchart}, in all questions, LIMEcraft was rated better than LIME. The bigger differences are visible in questions about how well the~superpixel boundaries were chosen and how would the~ability to change the~number of superpixels be rated.

The most appreciated functionalities provided by the~LIMEcraft tool were abilities to select own superpixels and interact with the~explanation method. Suprisingly, the~users disagreed on whether the~image modifications made it easier to detect model weaknesses. Also, opinions were divided on whether users would like to use LIMEcraft in the~future.

In examining the~correlation between responses, it was noted that the~greater the~familiarity with LIME, the~lower ratings users gave in response to the~questions: “How would you rate the~automatic superpixel division?”, and “How well were the~superpixel boundaries chosen?”. It might lead to the~conclusion that experienced LIME users have already discovered its weaknesses. Users more experienced in AI and LIME did not rate the~ability to catch model weaknesses by image modifications so highly.

The user study has shown that LIMEcraft is a~self-sufficient, end-to-end tool for model exploration. The functionality related to superpixel selection is especially valuable. The ability to edit the~image is less desirable than choosing own superpixels. The majority of people appreciated LIMEcraft under each of the~criteria studied.

\begin{figure}[h]
 \begin{tabular}{p{0.17\textwidth}p{0.17\textwidth}p{0.17\textwidth}p{0.15\textwidth}p{0.2\textwidth}}
 input image & LIMEcraft & LIME & Grad-CAM & Grad-CAM++\\
 \end{tabular}
    \includegraphics[width=\linewidth]{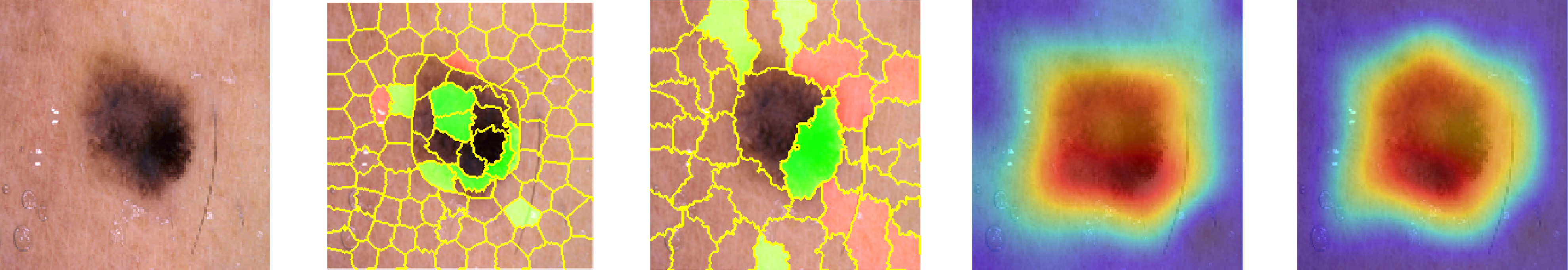} \\
    \includegraphics[width=\linewidth]{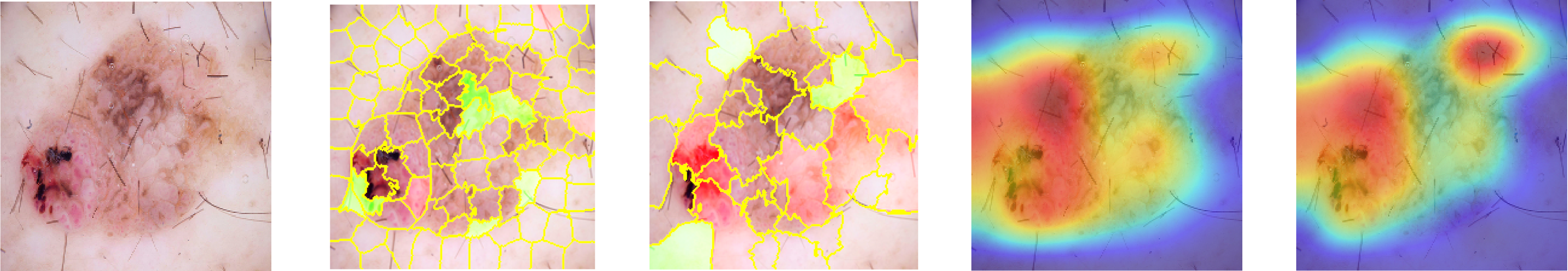} \\
    \includegraphics[width=\linewidth]{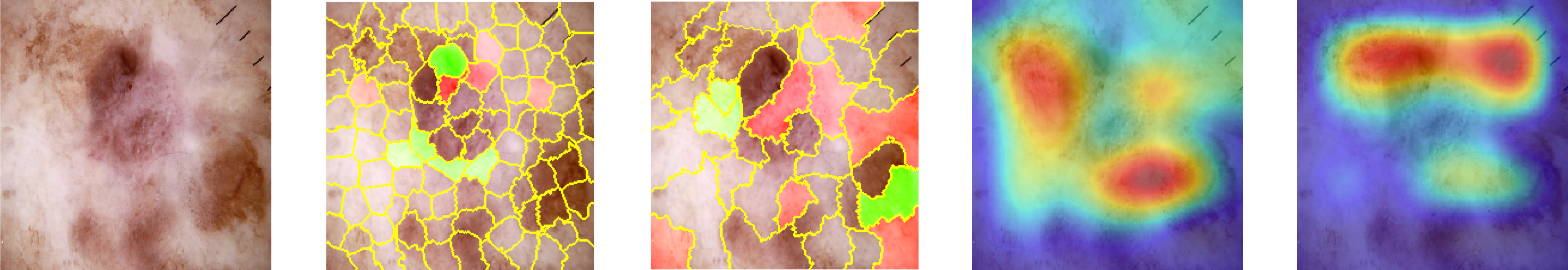}
    \caption{Model inference results for skin lesion classification using following explanation's algorithms: LIMEcraft, LIME \citep{Ribeiro2016}, Grad-CAM \citep{gradcam}, Grad-CAM++ \citep{gradcamplusplus}. In LIMEcraft and LIME, superpixels colored green indicate that this part of the~image has a~positive impact on prediction, while red parts have a~negative one. Grad-CAM and Grad-CAM++ are typically presented in “hot-to-cold” color scale. The red color shows the~areas that have the~most positive influence on model prediction, while blue presents the~most negative impact. The first row shows the~lesion with sharp borders. In the~second one, only the~wounded part was selected in LIMEcraft. The lesion in the~last row has blurred borders.} 
    
    \label{fig:xai_methods}
\end{figure}

\subsection{LIMEcraft versus other explanatory methods}

Various explanation algorithms are used to explain image models \citep{Samek}. Apart from explanations that work by perturbing input data, e.g., LIME and LIMEcraft, there are other popular methods that are based on gradients, e.g., Grad-CAM and Grad-CAM++.

Grad-CAM and Grad-CAM++ produce heatmaps that are calculated as a~result of the~features extracted from the~final convolutional layer of the~model. The heatmaps are coarse, as presented in Figure \ref{fig:xai_methods}. While comparing them to perturbation-based approaches, it is important to stress that they do not produce results with sharp boundaries of attention maps. Moreover, gradient-based algorithms show only areas where the~attention of a~model is present or absent, and do not take into account negative influence in predicting a~class.

\begin{table}[h]
\begin{tabular}{|p{0.5\textwidth}|p{0.5\textwidth}|}
\hline
\textbf{LIMEcraft}                                                                  & \textbf{LIME}                                                                               \\ \hline
Human-in-the-loop process                                                           & Fully automatic process                                                                     \\ \hline
Image data, audio data (transformed into a~spectrogram)                             & Image, text, tabular data                                                                   \\ \hline
Segmentation can be corrected                                                       & Segmentation cannot be corrected                                                            \\ \hline
Required masks (ready mask can be used)                                             & Not required masks                                                                          \\ \hline
Opportunity to focus explanations in a~specific area.                               & Lack of ability to focus explanations in a~specific area                                    \\ \hline
Opportunity to check what kind of image features contribute the~most in explanation & Lack of opportunity to check what kind of image features contribute the~most in explanation \\ \hline
More robust to noise                                                                & Less robust to noise                                                                        \\ \hline
Object of interest don’t have to have sharp boundaries                              & Object of interest should have sharp boundaries                                             \\ \hline
Opportunity to check how important are colored superpixels                          & Lack of opportunity to check how important are colored superpixels                          \\ \hline
\end{tabular}
\caption{Comparison of LIMEcraft and LIME.}
\label{tab:LIMEcraftvLIME}
\end{table}

Since LIMEcraft and LIME are both perturbation-based algorithms, it is worth noting the~differences between them. As presented in Table \ref{tab:LIMEcraftvLIME}, LIMEcraft offers more possibilities for exploring specific regions than LIME and is more robust to noise. However, it cannot be applied directly to text or tabular data. Taking a~human-in-the-loop of explanation makes LIMEcraft not work as an~automatic tool, but it provides the~possibility to add semantic meaning to the~explanation process. 

As shown in Figure \ref{fig:xai_methods}, in the~first row, the~sharp boundaries are easily identified by LIME. However, LIME divided the~lesion into only two parts. The superpixels in the~rest of the~image are much smaller, which means that the~most important part of the~image cannot be examined in detail. LIMEcraft offers a~solution to overcome this limitation. The ability to manipulate the~number of superpixels into which an~image is divided in selected region provides the~possibility to fine-tune the~size of the~superpixels. When we want small details, we divide the~part of the~image into more superpixels; whenever we want a~more global result, we divide it into fewer superpixels. 

In the~second row of Figure \ref{fig:xai_methods}, in comparison to LIMEcraft, LIME does not deal well with a~lesion of varying colors. In some places, the~lesion and the~unchanged skin are in the~same superpixel. The lack of semantic information in the~explained image is also shown in the~third row. 

Gradient-based methods versus perturbation-based methods are different approaches in explainability. LIMEcraft can be used to combine those two perspectives. The heatmap generated by gradient-based methods can be changed into masks and then used as ready input masks into LIMEcraft. It brings us a~closer look at the~hidden space of the~explained model.

\section{Conclusions}

To our knowledge, LIMEcraft is the~first LIME-based tool that allows the~user to be directly involved in the~construction of an~explanation. In this explanation tool, the~user can influence: (1) the~superpixel constructions of the~interpretable space on which LIME is based, (2) the~specific aspects of the~image to study how the~operation affects the~prediction of the~model and its explanation, (3) the~level of~detail of~the~obtained results. Moreover, operations such as perturbations of colors, shapes, and positions allow the~user to study the~sensitivity of the~model to changes.

The ability to select a~mask for LIMEcraft allows domain knowledge to be introduced into the~process of explaining. In addition, it makes it possible to apply the~method to cases where the~object under study is partially obscured or blends in color with its surroundings.
By checking the~relevance of each image feature, we can verify that the~model is not biased and that it has learned the~correct features, which will result in increasing the~model's safety and trustworthiness.

LIMEcraft might be applied to other inputs, such as audio data, because audio is often transformed into a~spectrogram that is an~image. Using LIMEcraft, rectangular-shaped superpixels can be created, which would allow the~audio to be divided by frequency. Also, the~concept of bringing human-in-the-loop can be applied to the~explanation process of tabular data, and humans can define the~thresholds for specific data transformations, e.g. grouping and discretization of variables.

Incorporating the~human into the~process of explainability allows to benefit from the~knowledge they have. However, if a~person does not expect an~element to be important, e.g., if they do not notice a~lesion, they will not mark it. For this reason, it is also important to be aware of the~risk of introducing bias into the~model. This is still an~open problem that needs further research.

We provided code and a~web application that can serve others to evaluate the~safety and robustness of their models.

\section*{Acknowledgment}

We would like to thank Dariusz Parzych, psychologist from Center for Innovation and Technology Transfer Management at Warsaw University of Technology, for a~valuable help in preparing and conducting user experience analysis, Piotr Piątyszek for helping in deployment of a~LIMEcraft web application, and 20 volunteers for participating in the~user study.

\section*{Statements and Declarations}
The research leading to these results received funding from NCN Sonata Bis-9 grant 2019/34/E/ST6/00052.

\bibliographystyle{unsrtnat}
\bibliography{references} 

\end{document}